\documentclass[conference]{IEEEtran}
\IEEEoverridecommandlockouts

\usepackage{lmodern}
\usepackage[T1]{fontenc}
\usepackage[utf8]{inputenc}
\usepackage[activate={true,nocompatibility},final,tracking=true,kerning=true,spacing=true,factor=1100,stretch=10,shrink=10]{microtype}
\usepackage{cite}
\usepackage{amsmath,amssymb,amsfonts}
\usepackage{algorithmic}
\usepackage{graphicx, wrapfig,lipsum}
\usepackage{textcomp}
\usepackage[inline]{enumitem}
\usepackage{xcolor}
\usepackage{url}
\usepackage{hyperref} 
\usepackage{subcaption} 
\usepackage{makecell}
\usepackage{fancyhdr}

\usepackage[belowskip=-15pt,aboveskip=0pt]{caption}
\def\BibTeX{{\rm B\kern-.05em{\sc i\kern-.025em b}\kern-.08em
    T\kern-.1667em\lower.7ex\hbox{E}\kern-.125emX}}

\title{Deep Slap Fingerprint Segmentation for Juveniles and Adults

\thanks{\textsuperscript{*} Authors contributed equally to this work.

\textsuperscript{**}  This material is based upon work supported by the Center for Identification Technology Research and the National Science Foundation (NSF) under Grant No.$650503$. Authors would also like to thank the \textit{Chameleon} cloud for providing the computational infrastructure required for this work \cite{keahey_lessons_2020}.
}

}
\author{\IEEEauthorblockN{M. G. Sarwar Murshed
\textsuperscript{*}, Robert Kline\textsuperscript{*}, Keivan Bahmani\textsuperscript{*}, Faraz Hussain, Stephanie Schuckers}\\
\IEEEauthorblockA{\textit{Electrical and Computer Engineering} \\
\textit{Clarkson University}, Potsdam, NY \\
{\tt \{murshem, kliner, bahmank, fhussain, sschucke\}@clarkson.edu}}
}

\begin{document}
\maketitle
\begin{abstract}
Many fingerprint recognition systems capture four fingerprints in one image. In such systems, the fingerprint processing pipeline must first segment each four-fingerprint slap into individual fingerprints. Note that most of the current fingerprint segmentation algorithms have been designed and evaluated using only adult fingerprint datasets. In this work, we have developed a human-annotated in-house dataset of 15790 slaps of which 9084 are adult samples and  6706 are samples drawn from children from ages 4 to 12. Subsequently, the dataset is used to evaluate the matching performance of the NFSEG, a slap fingerprint segmentation system developed by NIST, on slaps from adults and juvenile subjects. Our results reveal the lower performance of NFSEG on slaps from juvenile subjects. Finally, we utilized our novel dataset to develop the Mask-RCNN based Clarkson Fingerprint Segmentation (CFSEG). Our matching results using the Verifinger fingerprint matcher indicate that CFSEG outperforms NFSEG for both adults and juvenile slaps. The CFSEG model is publicly available at \url{https://github.com/keivanB/Clarkson_Finger_Segment}
\end{abstract}


\section{Introduction}
    

Due to their high accuracy and convenience, fingerprint-based recognition systems are now in widespread use, e.g. in border crossings, cell-phone authentication, and health care. Many fingerprint recognition systems use high-end multi-finger scanners instead of single finger scanners in order to achieve more accurate identification \cite{maltoni_handbook_2009}. In such systems, the fingerprint processing pipeline relies on a fingerprint segmentation model to localize individual fingerprints \cite{patriciaflanagannistgov_slap_2010} within each four-finger slap. As a result, slap fingerprint segmentation models are an integral part of fingerprint recognition systems as the loss of fingerprint ridge structure due to improper segmentation can significantly degrade the overall matching performance \cite{watson_slapsegii_2010}. 

To the best of our knowledge, all current slap fingerprint segmentation models have been developed using adult datasets \cite{patriciaflanagannistgov_slap_2010, Ko2010NBIS, gupta2018slap}.
However, since aging is known to affect the size and ridge-valley structure of the fingerprints, lower the quality and diminish the matching performance \cite{galbally_study_2018, haraksim_fingerprint_2019}, careful consideration and modeling are required to accurately segment and match slaps from juvenile subjects. However, the lack of publicly available juvenile datasets, as well as privacy concerns, have led to a dearth of work on this topic. 

In this work, we have developed a human-annotated in-house dataset of 15790 slaps (Children: 6706, Adult: 9084). Our new dataset allows us to accurately evaluate the performance of the NFSEG model on both adult and juvenile subjects. NFSEG is one of the most popular fingerprint segmentation system published by NIST \cite{Ko2010NBIS}. This segmentation system is used as a benchmark algorithm to evaluate the performance of a newly developed slap segmentation algorithm. Additionally, we utilized our human-annotated dataset to develop Clarkson Fingerprint SEGmentation (CFSEG) -- a novel deep learning-based fingerprint segmentation model designed to accurately segment both adult and juvenile slap fingerprints. Our test results using the \textit{Verifinger} fingerprint matcher (version 10, compliant with NIST MINEX \cite{watson2014fingerprint}), show that CFSEG outperforms NFSEG in both adult and juvenile cohorts of our dataset. 

\section{Collection and Annotation of Slap Fingerprints}
\label{sec:Dataset}
We combined several in-house adult and juvenile datasets to create a balanced and representative dataset for training and evaluating CFSEG. Our dataset contains a total of 15790 slap fingerprints (adults: 9084, juvenile: 6706) from 203 adults and 242 juvenile subjects. All slaps are captured at 500 ppi using the FBI certified \textit{Crossmatch L Scan Guardian (9000251)} fingerprint scanner \cite{noauthor_cross_nodate}. 

Initially, all slaps are segmented using the NFSEG model. Subsequently, we developed a Graphic User Interface (GUI) based on the open-source \verb+labelImg+ \cite{Tzutalin} platform for fingerprint annotation. Our GUI allows us to utilize NFSEG segmentations as a starting point for each capture. This process significantly reduced our annotation time to a practical time frame. Additionally, this pipeline allows us to utilize the estimated angle of rotation from the NFSEG and rotate each slap to the upright position. In the first stage, human annotators evaluated each slap for the correct angle of rotation. In the second stage, annotators cycle through slaps and correct the fingerprints that are incorrectly segmented by NFSEG while the rest of the bounding boxes remain untouched. This process resulted in a cleaned and annotated dataset with a total of 52674 localized fingerprints (adult: 30304, juvenile: 22370). Finally, we used a rigorous two stage 10-fold cross-validation process based on the unique identities in the dataset to derive our training, validation, and test sets. At each fold, $80\%$ of the adult and $80\%$ of the juvenile identities were used for training while the remaining $20\%$ of identities of both types were used for validation ($10\%$) and testing ($10\%$). 



\section{Clarkson Fingerprint Segmentation (CFSEG)}
Our CFSEG model for slap segmentation is based on the Mask-RCNN architecture \cite{he2017mask}. In this section, we discuss the overall Mask R-CNN model, the loss functions, and the training scheme used for training the CFSEG.

\subsection{Architecture of Mask R-CNN}

sh
Mask-RCNN is a simple and flexible two-stage deep architecture developed to perform semantic segmentation, object localization, and object instance segmentation of natural images \cite{he2017mask}. As a result, we adopt Mask-RCNN architecture for CFSEG. 

The first stage of this architecture contains the Region Proposal Network (RPN) which proposes candidate object bounding boxes. It consists of two networks, a Convolution Neural Network (CNN) and a Region Proposal Network (RPN). The CNN is the backbone of the Mask R-CNN architecture, and it is responsible for extracting feature maps from the input images. Any CNN model designed for image classification tasks (such as ResNet, MobileNet, or VGG) could be used as the backbone network \cite{Sebastian2019DeepBlueBerry}. Previous research has shown that the ResNet-FPN backbone provides better performance in terms of accuracy and speed on object detection tasks \cite{he2017mask}.  Therefore, we have used ResNet-101 \cite{he_deep_2015} as the backbone network in our experiments. On the other hand, the RPN is responsible for generating a set of `Region Proposals', which signify regions in the feature map that have a high probability of containing objects. In the second stage, a small, Fully Connected Neural Network (FCNN) takes the proposed regions from the first stage and predicts bounding boxes and object classes for each of them. Note that the proposed regions generated by the RPN can be of different sizes. However, fully connected layers in the networks only take a fixed size vector to make predictions. The ROIAlign \cite{he2017mask}, which is a modified version of Max-Pooling, is used to fix the size of these proposed regions. ROIAlign also fixes the misalignment which preserves the exact spatial location of an object and helps to improve overall accuracy.

A fully convolutions network is added (parallel with the existing branch for classification and bounding box regression) which is responsible for predicting segmentation masks in a pixel-to-pixel manner on each Region Of Interest (ROI). 

\subsection{Loss function}
For training RPNs, a class label indicating whether it is an object or not is assigned to each anchor. Anchor boxes are reference boxes placed at different positions in the input image. A positive label is assigned to an anchor if that anchor has an Intersection over Union (IoU) greater than 0.7 with any ground truth objects. A negative label is assigned to an anchor if that anchor has an IoU less than 0.3 with any ground truth objects. The anchors that have an IoU value between 0.3 and 0.7 are considered neutral and excluded from the training set. Shift and resizing operations are performed before training the RPN which helps make the anchor cover the ground truth object completely. With this definition, a multi-task loss function is used to train the Mask R-CNN. This loss function is divided into three parts which combine the loss of classification, localization, and segmentation as follows:
\vspace{-2pt}
\begin{equation}
    L = L_{cls} + L_{box} + L_{mask} \label{equ:Total_Loss}
\end{equation}

The class, bounding box, and mask losses are represented by $L_{cls}$, $L_{box}$ and $L_{mask}$ respectively. $L_{cls}$ is the log loss function over two classes carried out as a binary classification by predicting an object being a target object or not. The smooth L1 loss is used for bounding box loss $L_{box}$. Mask loss $L_{mask}$ is the average binary cross-entropy loss, and it is calculated for each class separately. These calculations prevent competition among the classes when generating masks. 
The details of different loss functions are as follows:
\begin{multline}
     L(\{p_i\}, \{t_i\}) = \frac{1}{N_{cls}} \sum_i L_{cls} (p_i, p_i^*) 
     + \frac{\lambda}{N_{box}} \\ \sum_i p_i^* L_{box}^{smooth}(t_i-t_i^*)
     + (-1* \frac{1}{m^2}) \sum_{1\leq i, j \leq m} [y_{ij} \log y_{ij}^k \\
     + (1 - y_{ij}) \log (1 - y_{ij}^k)] 
 \end{multline}

Here, $i$ represents the index number of an anchor and $p_i$ is the predicted probability of an anchor being an object. The ground-truth binary label is represented by $p_i^*$. $t_i$ and $t_i^*$ represent the vector with four coordinates of the predicted and ground-truth bounding box respectively. $m$ is the size of the mask output; $y_{ij}$ is the label of a cell (i,j) in the true mask for the region of size $m \times m$; $y^k_{ij}$ is the k-th mask predicted value of the same cell in the mask learned for the ground-truth class k.
\subsection{Training}
 
An image-centric training approached \cite{NIPS2015_14bfa6bb} is utilized in this work  and only positive anchors are used to determine the loss of the network. Our training approach is as follows:

\begin{itemize}
  \item N ROIs are generated from each image with the ratio of positive to negative ROIs of 1:3, where N is selected as 64 and 512 for backbone and FPN stages respectively.
  \item When training RPN, the anchors aspect ratio is 3, and the span is 5 scales. The RPN is trained individually and does not share features with other parts of the Mask-RCNN network. 
  \item The Stochastic Gradient Descent (SGD) with a learning rate of 0.001, momentum of 0.9, and decay of 0.0001 are used to train the network.
\end{itemize}

During inference, the region proposal number increases to 300 and 1000 for the backbone and FPN stages, respectively. This increment helps the network not miss any potential regions during inference time. The bounding box prediction branch considers each proposal proposed by the backbone and FPN network. Then the non-maximum suppression is performed to remove the low-scoring ROIs. To keep computational overhead to a minimum during training and inference, the mask branch only runs over the 50 highest scoring ROIs. 


\begin{figure}[t]
  \centering
  \includegraphics[width=0.8\linewidth]{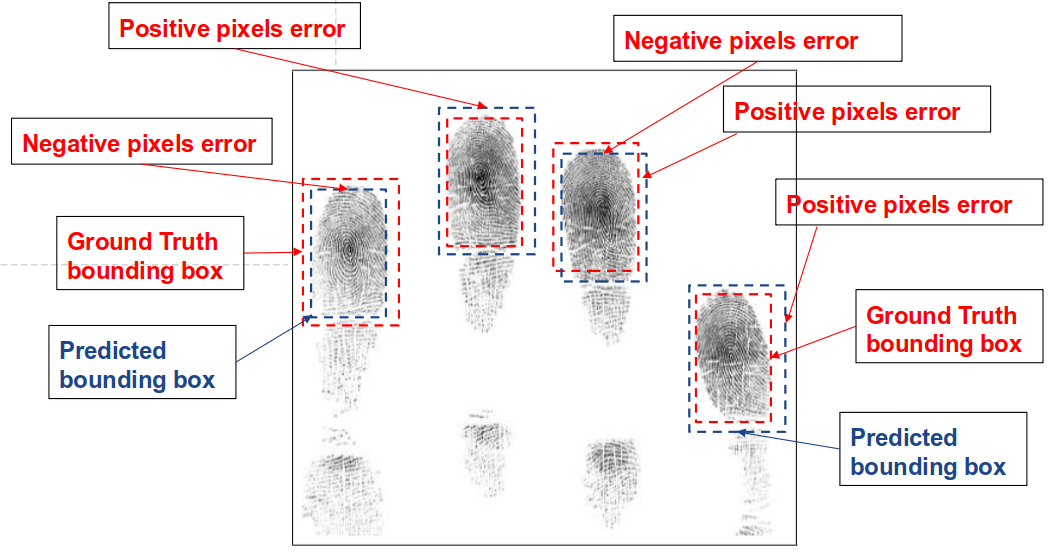}
\caption{Example for calculating positive and negative errors between the predicted and ground truth bounding box.}
\label{fig:PixelError}
\end{figure}

\section{Evaluation and Results}

In this work, we utilized both the Mean Absolute Error (MAE) of the predicted bounding boxes and fingerprint matching performance as our evaluation metrics. In this section, we describe our evaluation process and results. 

\subsection{Mean Absolute Error (MAE)}
The MAE measures the distance between the detected bounding box and the annotated ground-truth bounding box in terms of pixels. Slap fingerprint segmentation models have to find the balance between over-extending the bounding boxes (over-segmentation) vs. excessively reducing the predicted bounding box (under-segmentation). Over-segmentation can lead to the ridge-valley structure of other fingerprints leaking into the bounding box of other fingerprints. This extra noise can potentially degrade the matching performance. On the other hand, under-segmentation can cause the loss of the valuable parts of the fingerprints. 

Previous work has identified that under-segmentation beyond 32 pixels from the sides and 64 pixels from the top or bottom of the fingerprint can negatively affect the matching performance \cite{watson_slapsegii_2010}. As a result, we calculate and report MAE for \emph{each side} of the bounding box individually, i.e. the error is calculated by measuring how far those four sides are from the four sides of the ground truth bounding box. In \autoref{fig:PixelError}, the red dotted rectangle represents the ground-truth bounding box generated by a human-annotator and the blue rectangle represents the detected bounding box by a fingerprint detection model. If any side of a detected bounding box captures more information than the side of the ground-truth bounding box, we consider it as a positive error. On the other hand, if any side of a detected bounding box captures less information than the ground-truth bounding box, we consider it as a negative error. Finally, the MAE for each side is calculated using the equation \ref{equ:MAEs}, Where, $n$ is the total number of fingerprints in the test dataset. $X$ represents any side such as left, right, top, bottom, of the bounding box. 
\begin{equation}
\label{equ:MAEs}
    MAE = \frac{\sum_{i = 0} ^ {n} abs(X \: error_i)}{Total\_fingerprints\_in\_the\_dataset}
\end{equation}

\begin{figure*}[ht]
\begin{center}
\begin{subfigure}{0.45\textwidth}
    \includegraphics[width=.7\linewidth]{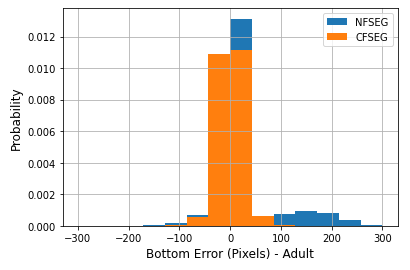}
\end{subfigure}
\begin{subfigure}{0.45\textwidth}
    \includegraphics[width=.7\linewidth]{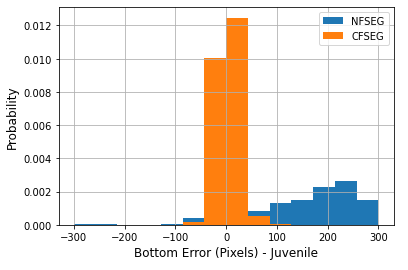}
\end{subfigure}
\caption{The histogram of the bottom error (pixels) from predicted bounding boxes by the NFSEG and CFSEG models for adult (Left) and Juvenile (Right) subjects.}
\label{fig:fullComChart}
\end{center}
\end{figure*}

\begin{figure}[h]
  \centering
  \includegraphics[width=.5\linewidth]{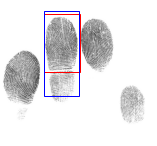}
\caption{Example of over-segmentation error in NFSEG. The red rectangle shows the ground-truth bounding box while the blue rectangle shows the detected bounding box by the NFSEG.}
\label{fig:pixelLossExample}
\end{figure}

Table \ref{table:AvgPixelLoss} shows the mean and standard deviation of the MAE for NFSEG and CFSEG models evaluated using our dataset. We can observe that, compared to NSFEG, CFSEG has lower MAE in almost all directions and produces more precise bounding boxes for both adults and juvenile subjects. Additionally, our results confirm that compared to adults, NFSEG suffers from higher MAE in all directions in the juvenile cohort of our dataset. Furthermore, we observe that in both adult and juvenile subjects, NFSEG has difficulty in accurately localizing the lower boundary of fingerprints. Figure \ref{fig:fullComChart} illustrates the histogram of the bottom errors for adult and juvenile subjects. We can observe that while CFSEG maintains the performance in Juvenile subjects, NFSEG is particularly susceptible to this type of error in juvenile subjects. Our failure analysis on such cases revealed that the high error observed in this direction is because of the over-segmentation at the distal interphalangeal joint. Figure \ref{fig:pixelLossExample} depicts an example of this problem, where the red bounding box indicates the ground truth and the blue bounding box presents the bounding box predicted by NFSEG. Additionally, the long tail of the error histogram of the NFSEG model for the juvenile subjects suggests that NFSEG over-segmented many juvenile samples below the Minimum Tolerance Limit (MTL) of -64 pixels suggested by Slapseg-II \cite{watson_slapsegii_2010}. This can potentially reduce the matching performance of the system.   




\begin{table} [h]
    \renewcommand{\arraystretch}{1}
\caption{\small Mean Absolute Error (MAE) [Mean (Std.)] for NFSEG and CFSEG}
\label{table:AvgPixelLoss} 
  \centering
  \begin{tabular}{|p{0.85cm}|p{1.4cm}|p{1.44cm}|p{1.65cm}|p{1.45cm}|}
    \hline 
\makecell{Age \\ Group} & \multicolumn{2}{|c|}{Adults} & \multicolumn{2}{|c|}{Children}  \\    
\hline 
Model & NFSEG & CFSEG & NFSEG & CFSEG \\
\hline
\hline 
Left  & 07.13(28.59) & 08.21(14.66) & 12.09(38.19) & 08.36(16.25) \\
\hline
Top  & 13.18(44.70) & 13.05(20.26) & 28.47(92.51) & 13.77(21.13) \\
\hline
Right  & 09.46(31.52) & 07.73(13.78) & 13.60(34.55) & 07.21(14.16) \\
\hline
Bottom  & 35.23(84.73) & 16.60(24.41) & 102.87(159.00) & 15.00(22.92) \\
\hline
\end{tabular}
\vspace{-28pt}
\end{table}


\subsection{Fingerprint Matching}
In the second stage of the evaluation process, we utilized the \textit{Verifinger} fingerprint matcher (version 10, compliant with NIST MINEX \cite{watson2014fingerprint}) to compare the matching accuracy using the fingerprints segmented with NFSEG and CFSEG. This helps us better analyze the effects of segmentation accuracy on the overall performance of the fingerprint recognition system. In order to evaluate the matching accuracy, we used the annotated ground truth, NFSEG, and CFSEG bounding boxes. For every fingerprint in the dataset, we evaluated all the mated comparisons for our genuine distribution (172348), while randomly selecting 20 non-mated fingerprints to construct an imposter distribution (441322 imposter comparisons). In total, we performed $~600,000$ comparisons using all 10 fingers. 

Our results indicate that for both adults and juvenile cohorts the fingerprints segmented with CFSEG provide higher accuracy across operation points. Additionally, we observe lower overall matching performance in juvenile subjects compared to adults. Finally, Table \ref{table:Matching} depicts the True Positive Rate (TPR) at False Positive Rate (FPR) of 0.1\%  for fingerprints segmented using NFSEG, CFSEG, and human-annotators. 



\begin{table}[h]
    \renewcommand{\arraystretch}{1.5}
\caption{\small TPR [Mean (Std.)] at FPR of 0.001 for NFSEG, CFSEG, and Ground Truth.}
\label{table:Matching} 
  \centering
  \begin{tabular}{|c|c|c|}
    \hline 
{Model}  & Adults & Children\\ 
\hline 
NFSEG & 0.9972 (0.0027)  & 0.9675 (0.0135)  \\
\hline
CFSEG & 0.9977 (0.0026)  & 0.9687 (0.0135)  \\
\hline
Ground-Truth & 0.9991 (0.0011)   & 0.9716 (0.0137) \\
\hline
\end{tabular}
\vspace{-28pt}
\end{table}


\section{Conclusion and Limitations}
The experimental results indicate that our novel CFSEG model provides more precise bounding box predictions with respect to NFSEG for both adult and juvenile subjects. However, CFSEG currently depends on the estimated rotation angle of the slap acquired from NFSEG. This will be addressed in future work by the introduction of a regression sub-network for angle prediction and artificially augmenting the training dataset with rotated slaps. Additionally, our fingerprint matching experiment is conducted using the human-annotated class labels (i.e. index, middle, or ring fingers) for all algorithms to produce a meaningful comparison between the predicted bounding boxes. Future work can focus on designing a class prediction sub-network to integrate this task into the CFSEG and further improve the overall performance. Finally, the trained CFSEG model is publicly available to other researchers at \url{https://github.com/keivanB/Clarkson_Finger_Segment}.




\bibliographystyle{IEEEtran}
\bibliography{IEEEabrv, references}

\end{document}